\documentclass[conference]{IEEEtran}
\IEEEoverridecommandlockouts
\usepackage{cite}
\usepackage{amsmath,amssymb,amsfonts}
\usepackage{algorithmic}
\usepackage{graphicx}
\usepackage{textcomp}
\usepackage{xcolor}
\usepackage{ragged2e} 
\usepackage{booktabs,makecell, multirow, tabularx}
\def\BibTeX{{\rm B\kern-.05em{\sc i\kern-.025em b}\kern-.08em
    T\kern-.1667em\lower.7ex\hbox{E}\kern-.125emX}}
\begin{document}

\title{Vision-based Vehicle Re-identification in Bridge Scenario using Flock Similarity
}

\author{\IEEEauthorblockN{Chunfeng Zhang}
\IEEEauthorblockA{\textit{School of Intelligent Systems Engineering} \\
\textit{Sun Yat-sen University}\\
Shenzhen, China \\
zhangchf25@mail2.sysu.edu.cn}
\and
\IEEEauthorblockN{Ping Wang*}
\IEEEauthorblockA{\textit{School of Intelligent Systems Engineering} \\
\textit{Sun Yat-sen University}\\
Shenzhen, China \\
wangp358@mail.sysu.edu.cn}
}

\maketitle
\begin{abstract}
Due to the needs of road traffic flow monitoring and public safety management, video surveillance cameras are widely distributed in urban roads. However, the information captured directly by each camera is siloed, making it difficult to use it effectively. Vehicle re-identification refers to finding a vehicle that appears under one camera in another camera. This technology can correlate the information captured by multiple cameras, which is of great significance for realizing the intelligence of the monitoring system. While license plate recognition plays an important role in some applications, there are some scenarios where re-identification method based on vehicle appearance are more suitable. The main challenge is that the data of vehicle appearance has the characteristics of high inter-class similarity and large intra-class differences. Therefore, it is difficult to accurately distinguish between different vehicles by relying only on vehicle appearance information. At this time, it is often necessary to introduce some extra information, such as spatio-temporal information. Nevertheless, the relative position of the vehicles rarely changes when passing through two adjacent cameras in the bridge scenario. In this paper, we present a vehicle re-identification method based on flock similarity, which improves the accuracy of vehicle re-identification by utilizing vehicle information adjacent to the target vehicle. When the relative position of the vehicles remains unchanged and flock size is appropriate, we obtain an average relative improvement of 204\% on VeRi dataset in our experiments. Then, the effect of the magnitude of the relative position change of the vehicles as they pass through two cameras is discussed. We present two metrics that can be used to quantify the difference and establish a connection between them. Although this assumption is based on the bridge scenario, it is often true in other scenarios due to driving safety and camera location.
\end{abstract}


\section{Introduction}
With the acceleration of urbanization and the growth of traffic flow, the importance of vehicle re-identification technology in intelligent transportation systems has become increasingly prominent. This technology is often used to identify and track the same vehicle under different surveillance cameras, which helps to improve the efficiency of traffic supervision. Due to the reduction of camera costs and the maturity of video compression technology, distributed cameras have been widely deployed in urban roads. The huge video surveillance system plays an important role in monitoring road traffic flow and public safety management. However, the information captured directly by each camera is siloed, making it difficult to use it effectively. When anomalies are found, it is necessary to manually look up different cameras to obtain more comprehensive information. In large-scale surveillance systems, manual search takes a lot of time and effort. Therefore, the use of artificial intelligence method to carry out real-time correlation and analysis of the data captured by the cameras can achieve more efficient and accurate management and monitoring. Vehicles are the main participants in urban roads, so vehicle re-identification technology is one of the most important technologies to integrate the information captured by different cameras.

In some scenarios, re-identification based on vehicle appearance may be more suitable than license plate recognition. Although the license plate recognition plays an important role in some specific occasions, there are some situations where it is not available. The images captured by the camera may have low resolution due to long distance, high vehicle speed, poor lighting, etc. In addition, improper shooting angles and occlusion problems may also result in a failure of license plate recognition. In contrast, the entire vehicle appearance area is relatively large, so vehicle re-identification based on vehicle appearance has relatively low requirements for image resolution. At the same time, if an object blocks the license plate, then it does not necessarily block the entire vehicle. Therefore, vehicle re-identification based on vehicle appearance is more robust to occlusion problems. Additionally, it has better protection of privacy.

\begin{figure}[t]
	\centering
	\includegraphics[width=0.3\textwidth]{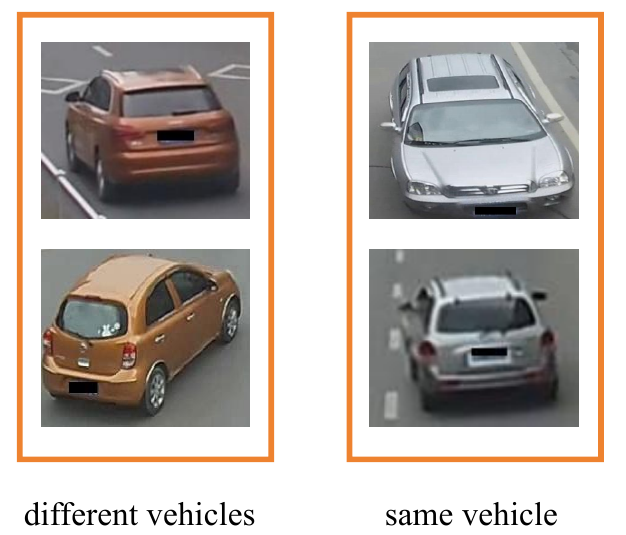}
	\caption{Some examples from VeRi dataset. \textbf{Left:} different vehicles with a highly similar appearance.
		\textbf{Right:} the same vehicle from different viewpoints.}\label{fig1}
\end{figure}

At present, vehicle re-identification method based on vehicle appearance faces many challenges, such as low image resolution, occlusion, different viewpoint, and changes in lighting conditions. In addition, the data of vehicle appearance has the characteristics of high inter-class similarity and large intra-class differences. As shown in Fig.~\ref{fig1}, due to the limited color and model of vehicles, different vehicles may have a highly similar appearance. In addition, due to the different shooting angles, the images of the same vehicle at different times, different places, and different angles show different appearance characteristics~\cite{r4}. This further increases the difficulty of vehicle re-identification based on vehicle appearance.

To address these challenges, previous researchers have proposed a number of approaches based on neural network models, which have proven to be powerful in feature extraction and perform well in many computer vision tasks. The papers of~\cite{r2,r11,r7,r9} propose some new network architectures and loss functions in order to learn more robust features. In this respect, vehicle re-identification is similar to person re-identification, where researchers have made much progress~\cite{r13,r14,r15,r16}. Furthermore, it is a popular method to improve the accuracy of vehicle re-identification by introducing extra information, such as orientation~\cite{r4}, color, vehicle model~\cite{r1}, spatiotemporal information~\cite{r3,r8}, etc. Re-ranking is a simple but very effective post-processing step that is often applied in re-identification tasks to improve sorting results~\cite{r4,r1,r3}. Another route is to avoid the use of complex frameworks and additional labels to balance the requirements of accuracy and computational efficiency in large-scale vehicle re-identification tasks~\cite{r5,r6}.

\begin{figure}[t]
	\centering
	\includegraphics[width=0.45\textwidth]{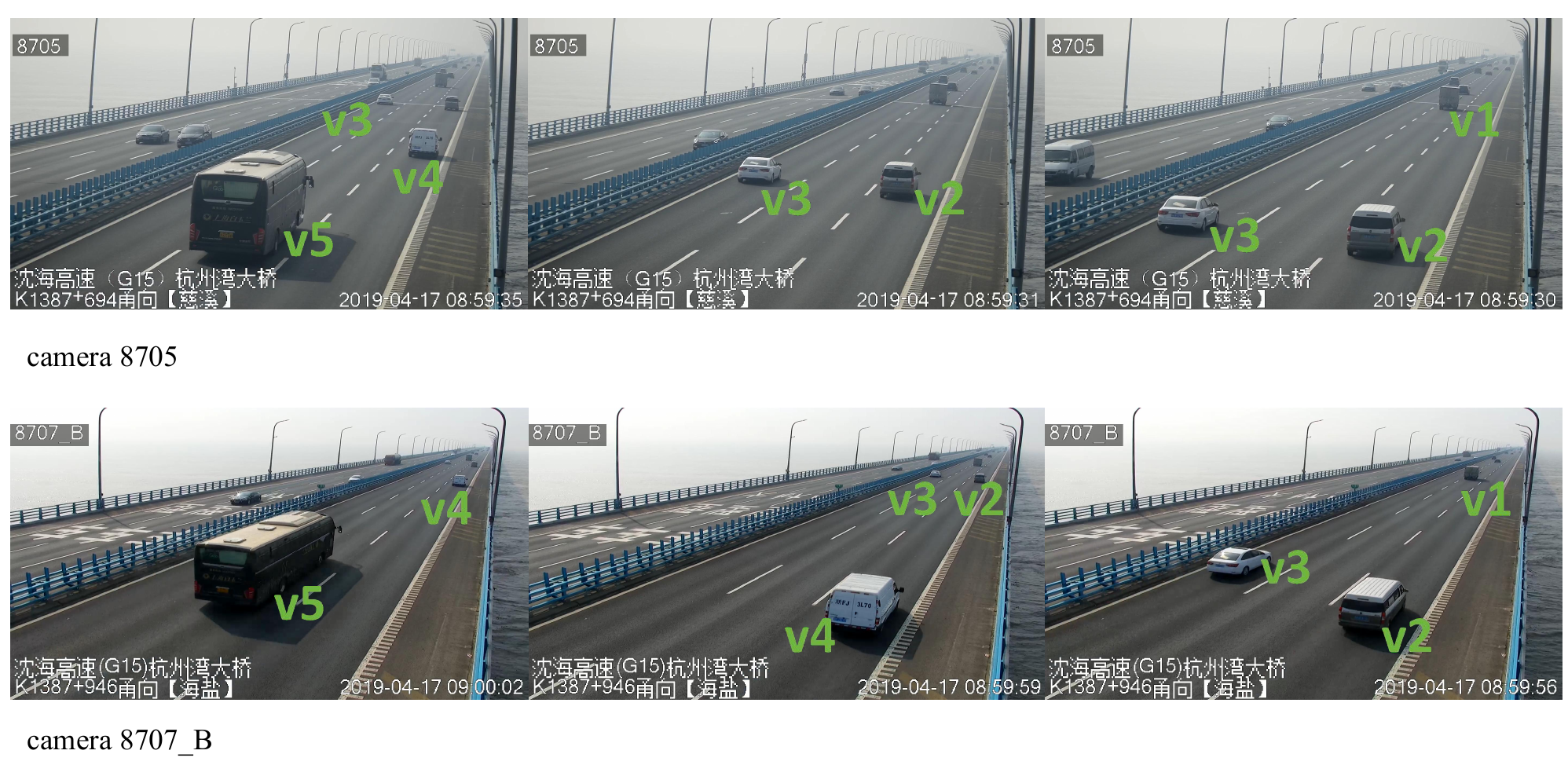}
	\caption{Scene of some vehicles passing through two cameras in the bridge scenario. The relative position between them rarely changes.}\label{fig2}
\end{figure}

Previous work has focused on extracting features from a single vehicle sample through neural networks to calculate the similarity between different vehicle samples. However, the data of vehicle appearance has the characteristics of high inter-class similarity and large intra-class differences, so it is often difficult to distinguish them by the extracted features when two samples are highly similar. In some cases, when vehicles pass through two adjacent cameras, the relative position between them rarely changes, we can distinguish it by the vehicle adjacent to the target vehicle. As shown in Fig.~\ref{fig2}, the relative position between them rarely changes when passing through two adjacent cameras in the bridge scenario. There are both subjective and objective reasons for this. On the one hand, we can improve the safety of driving by following a vehicle and keeping a safe distance from it. Frequent lane changing and overtaking require other vehicles to make corresponding evasions or adjustments, which will increase the risk of accidents. Therefore, overtaking is even prohibited in some special sections. On the other hand, cameras are often installed at intersections or vehicle diversions to monitor traffic flow. As a result, the distance between two adjacent cameras tends to be relatively short, which means that the time it takes for a vehicle to pass through the road between two adjacent cameras is generally not too long. Consequently, even if some vehicles change lanes to overtake, it is unlikely that the relative positions between vehicles will change much in a short period of time. In a nutshell, it is reasonable to assume that the relative position between vehicles rarely changes when they pass through two adjacent cameras. This assumption holds true not only for the bridge scenario, but also for many others~\cite{r18}. In this work, we propose a vehicle re-identification method based on flock similarity and demonstrate its effectiveness through experiments.

\section{Flock Similarity}

\begin{figure}[t]
	\centering
	\includegraphics[width=0.45\textwidth]{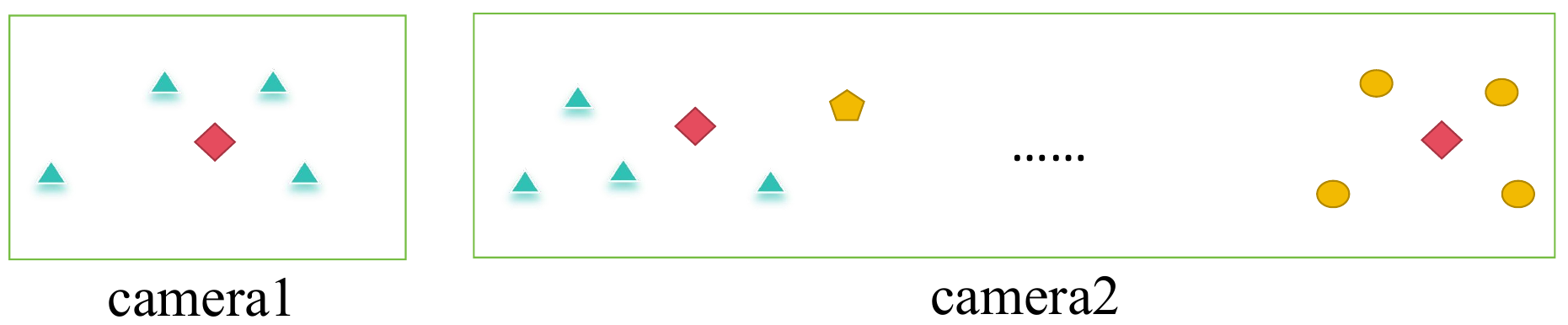}
	\caption{A simple schematic representation of vehicles passing through two cameras. }\label{fig3}
\end{figure}

\begin{figure}[t]
	\centering
	\includegraphics[width=0.45\textwidth]{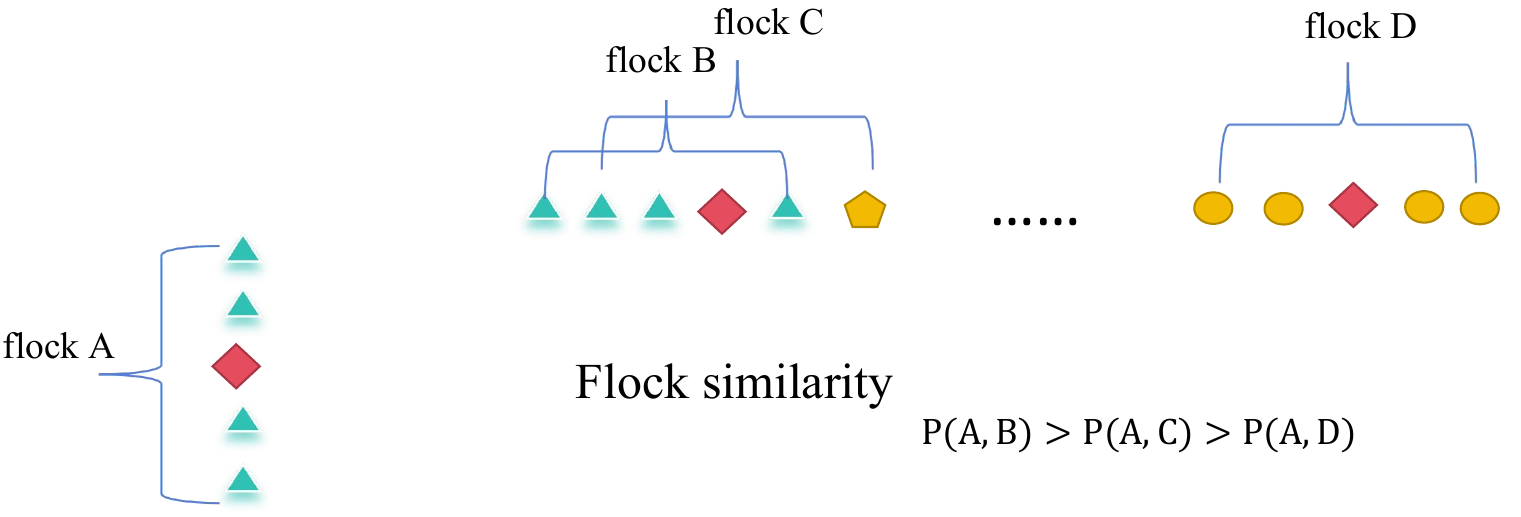}
	\caption{A simple example of flock similarity.}\label{fig4}
\end{figure}

The relative position between vehicles rarely changes when passing through two adjacent cameras, so vehicles located near the target vehicle under Camera1 will also be located near the target vehicle under Camera2, as shown in Fig.~\ref{fig3}.
Suppose that the diamond shape under Camera1 represents the target vehicle to be re-identified, and there are two individuals under Camera2 that are highly similar to it. In this case, it is difficult to distinguish them based on individual similarity. When using flock similarity, the problem becomes very simple. As shown in Fig.~\ref{fig4}, assuming the selected flock size is five, flock $A$ is a set composed of the target vehicle and four vehicles adjacent to the target vehicle under Camera1. We calculate the similarity between flock $A$ and each flock under Camera2 to find the flock with the highest similarity to the target flock. It is not difficult to find that flock $B$ have the highest similarity to flock $A$, followed by flock $C$, and flock $D$ has the lowest similarity to flock $A$. Although flock $D$ contains an individual highly similar to the target vehicle, its similarity with the target flock (flock $A$) is quite low.

For two flocks of the same size, flock $A= \{a_1, a_2, a_3,\ldots, a_n\}$, flock $B=\{b_1, b_2, b_3,\ldots, b_n\}$, let $p_{ij}$ represent the similarity between individual $a_i$ and individual $b_j$. We define the similarity between flock $A$ and flock $B$ as:
{\setlength\arraycolsep{2pt}
\begin{align}
	\begin{split}
		P(A,B) =
		& \max\{\frac{\sum_{i=1}^{n}\sum_{j=1}^{n}p_{ij}x_{ij}}{n}|\\
		& {}\sum_{j=1}^{n}x_{ij} = 1,i=1,2,\ldots,n;\\
		& {}\sum_{i=1}^{n}x_{ij} = 1,j=1,2,\ldots,n;\\
		& {}x_{ij}\in\{0,1\},i,j=1,2,\ldots,n{}\},
	\end{split}
\end{align}
and then the flock similarity has the following properties:
\begin{enumerate}
	\item The value range of flock similarity is 0--1.
	\item Symmetry, i.e. $P(A,B)=P(B,A)$.
	\item When the flock size is 1, flock similarity degenerates into individual similarity.
	\item The relative position of individuals within a flock has no effect on flock similarity.
	\item On the premise that the individual similarity calculation is ideal (i.e., for the same individual, their similarity is 1), the similarity between two flocks made up of the same individuals is 1.
\end{enumerate}

According to property 3), individual similarity is a special case of flock similarity. When the flock size is 1, that is, there is only one individual in the flock, the similarity between two flocks is equal to the similarity between two individuals. According to property 4), when vehicles pass through two adjacent cameras, the relative position between them slightly changes, and the method based on flock similarity is still applicable. In the process of calculating flock similarity, vehicles between two flocks are actually matched one by one. In general, we choose the matching result of the flock centered on the target vehicle as the matching result of the target vehicle.

When calculating the flock similarity, the elements in the set are finite (it is not greater than $n!$), so we can use the enumeration method to find all the $\frac{\sum_{i=1}^{n}\sum_{j=1}^{n}p_{ij}x_{ij}}{n}$ that meet the conditions, and then compare them one by one, and finally find the maximum value in the set (i.e., the similarity between flock $A$ and flock $B$), but its time complexity is $O(n!)$. This approach does not work when the value of $n$ is large, so we need to continue the discussion on how to find the maximum value effectively.

The calculation of flock similarity can be transformed into the solution of the following integer programming problem:
\begin{equation}
	\max z=\frac{\sum_{i=1}^{n}\sum_{j=1}^{n}p_{ij}x_{ij}}{n}\label{eq2}
\end{equation}
\begin{displaymath}
	s.t. \left\{ \begin{array}{ll}
		\sum_{j=1}^{n}x_{ij} = 1, & \textrm{$i=1,2,\ldots,n$}\\
		\sum_{i=1}^{n}x_{ij} = 1, & \textrm{$j=1,2,\ldots,n$}\\
		x_{ij}\in\{0,1\}, & \textrm{$i,j=1,2,\ldots,n$}
	\end{array} \right.
\end{displaymath}
Let 
\begin{equation}
	b_{ij} = 1 - p_{ij}
\end{equation}
After transformation, it can be transformed into the solution of following objective function
\begin{equation}
	\min z’=\frac{\sum_{i=1}^{n}\sum_{j=1}^{n}b_{ij}x_{ij}}{n}
\end{equation}
The problem has now been transformed into a minimization assignment problem. $b\geq 0$, meets the conditions of the Hungarian algorithm, which can be used to find the optimal solution to the minimization assignment problem within $O (n ^ 3)$ time complexity.
The minimum solution obtained is the maximum solution to the original problem.
\begin{align}
	\begin{split}
	\sum_{i=1}^{n}\sum_{j=1}^{n}b_{ij}x_{ij}
	& = \sum_{i=1}^{n}\sum_{j=1}^{n}(1-p_{ij})x_{ij} \\
	& = \sum_{i=1}^{n}\sum_{j=1}^{n}x_{ij} -\sum_{i=1}^{n}\sum_{j=1}^{n}p_{ij}x_{ij} \\
	& = n-\sum_{i=1}^{n}\sum_{j=1}^{n}p_{ij}x_{ij}
		\end{split}
\end{align}
Since $n$ is a constant, when $\sum_{i=1}^{n}\sum_{j=1}^{n}b_{ij}x_{ij}$ is minimized, $\sum_{i=1}^{n}\sum_{j=1}^{n}p_{ij}x_{ij}$ is maximized.

\section{Prepare for experiments}
To make the experiments simpler, we use a Siamese Neural Network (SNN) with VGG16~\cite{r12} as the backbone feature extraction network to calculate the similarity between the two individuals. SNN is a special type of neural network that is used to determine whether two inputs are similar or identical. This network has two inputs, denoted as $x1$ and $x2$. Each input is feature extracted through a separate neural network, and the outputs of the two neural networks are then combined to calculate the loss. These two neural networks can be exactly the same or different. Each neural network maps the input to a new feature space, forming a representation of the input in the new space. In the vehicle re-identification task, since the types of the two inputs are the same, we use two identical neural networks for feature extraction, that is, the weights are shared between the two neural networks, as shown in Fig.~\ref{fig5}.
\begin{figure}[t]
	\centering
	\includegraphics[width=0.35\textwidth]{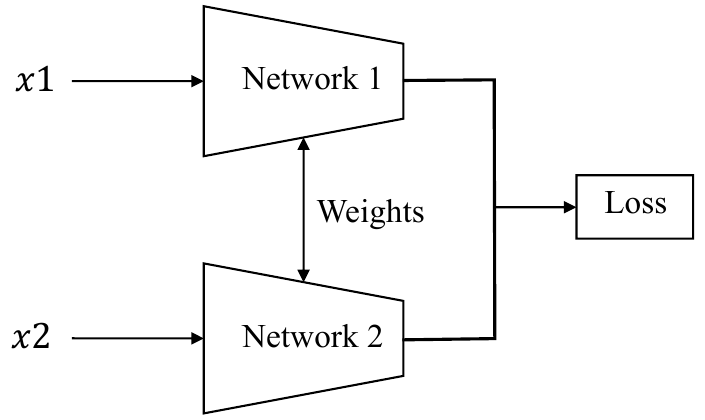}
	\caption{The Siamese Neural Network architecture.}\label{fig5}
\end{figure}
VGG16 is a convolutional neural network model with a relatively simple and easy-to-understand model structure, which mainly consists of 13 convolutional layers and 3 fully connected layers, without using any complex tricks or modules. This network structure is not only easy to understand, but also effective in learning the similarity relationship between two inputs.

The training data we used was VeRi~\cite{r10}, a public dataset commonly used in vehicle re-identification tasks. It contains more than 50,000 images of 776 vehicles taken by 20 cameras, which are captured in unconstrained surveillance scenes in the real world, each vehicle is taken by 2--18 cameras with different viewing angles, lighting, resolution and occlusion. In addition, it is marked with sufficient license plates and spatio-temporal information. In this paper, we only use vehicle image data from the VeRi dataset.

\section{Experiments}
Vehicle re-identification based on flock similarity requires searching in an ordered image list to find the same vehicle captured by another camera. The vehicle images in the list are sorted according to their order of appearance in the camera. In order to keep simple and focus on the impact of different flock sizes on the results of the experiment (individual can be seen as a flock of size 1), we consider such a simple scenario. Suppose there is $N$ vehicles driving past two cameras (Camera1 and Camera2), and $x_i$ represents the order in which $vehicle_i$ appears in Camera1, where $x_i\in \{0,1,\ldots,N-1\}$ and if $i\neq j$, then $x_i\neq x_j$. In the same way, $y_i$ represents the order in which $vehicle_i$ appears in Camera2, where $y_i\in \{0,1,\ldots,N-1\}$ and if $i\neq j$, then $y_i\neq y_j$. Therefore, we can use two vectors, $\boldsymbol {x}$ and $\boldsymbol {y}$, to represent the order in which the vehicles pass through the camera. $x_i=j$ represents that the order in which $vehicle_i$ appears under Camera1 is $j$.  In the same way, $y_i=j$ represents that the order in which $vehicle_i$ appears under Camera2 is $j$. Without loss of generality, $x_i$ can be set to $i$, $i\in \{0,1,\ldots,N-1\}$. In the experiment, we used different methods to find the corresponding $y_i$ for each $x_i$, and used accuracy as the evaluation index of them. The result can be represented by the vector $\boldsymbol{ \hat{y}}$, and we avoid any post-processing of the result, so the set $\{\hat{y}_i|i\in \{0,1,\ldots,N-1\}\}$ is a subset of the set $\{y_i|i\in \{0,1,\ldots,N-1\}\}$. 

Although the two vectors are of the same length and contain the same elements, it does not affect the scientific nature of the experiment. This is because each query is made on an individual basis. We find the corresponding $y_i$ for each $x_i$ just to calculate the accuracy. In addition, the flock size is much smaller than the length of vector y, i.e., for each flock, vector $\boldsymbol{y}$ still contains many elements outside the flock.

We use the images from VeRi test set to construct two ordered lists of images (List1 and List2). Two images taken by different cameras for each vehicle are placed in the same position in the two lists. List1 represents the vehicle images captured by Camera1, while List2 represents the vehicle images captured by Camera2. Vehicles appear in the same order under both cameras, so $y_i$=$x_i$=$i$, $i\in \{0,1,\ldots,N-1\}$, which means that the relative position of the vehicles is completely unchanged when the vehicles pass through two cameras. We construct lists with lengths of N=50, 100, 200, respectively, the results of the experiment are shown in Table~\ref{tab1}.
\begin{table}[htbp]
	\caption{Evaluation results when the relative position remains unchanged.}\label{tab1}
	\begin{center}
		\begin{tabular}{|c|c|c|c|c|}
			\hline
			\multicolumn{2}{|c|}{\multirow{2}{*}{Rank-1 accuracy}}&\multicolumn{3}{c|}{The length of the list} \\
			\cline{3-5} 
			\multicolumn{2}{|c|}{} &\textit{N=50}&\textit{N=100}& \textit{N=200}\\
			\hline
			\multirow{5}{*}{Flock size}& 1 & 0.44 & 0.31 & 0.22\\
			
			& 3 & \textbf{0.92} & 0.84 & 0.825 \\
			
			& 5 & \textbf{0.92} & \textbf{0.9} & \textbf{0.91} \\
			
			& 7 & 0.9 & 0.87 & 0.905 \\
			
			& 9 & 0.84 & 0.83 & 0.875 \\
			
			\hline
		\end{tabular}
	\end{center}
\end{table}

It can be seen that when the vehicles pass through two cameras and the relative position between them is completely unchanged, the accuracy of the method based on flock similarity is much higher than the method based on individual similarity (flock size$=$1). In addition, the method based on flock similarity shows better robustness to the change of list length. When the list length increases, the accuracy of the method based on flock similarity decreases relatively slowly. Because as the length of the list increases, the probability of two or more similar flocks appearing in the list is much smaller than the probability of two or more similar individuals appearing in the list.
\begin{figure*}[t]
	\centering
	\includegraphics[width=0.85\textwidth]{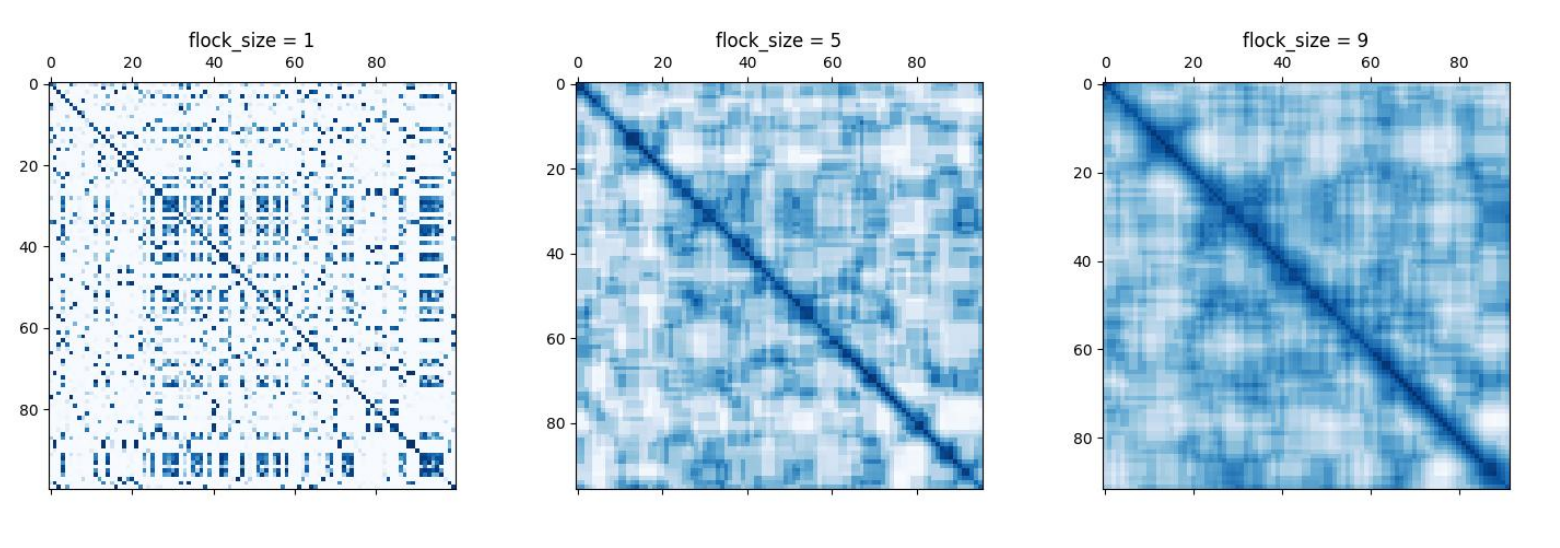}
	\caption{Flock similarity visualizations. The similarity between flocks of vehicles captured by two cameras when the relative position remains unchanged. \textbf{Left:} the size of flock is 1 (individual). \textbf{Middle:} the size of flock is 3.
		\textbf{Right:} the size of flock is 5.}\label{fig6}
\end{figure*}
\begin{figure*}[t]
	\centering
	\includegraphics[width=0.85\textwidth]{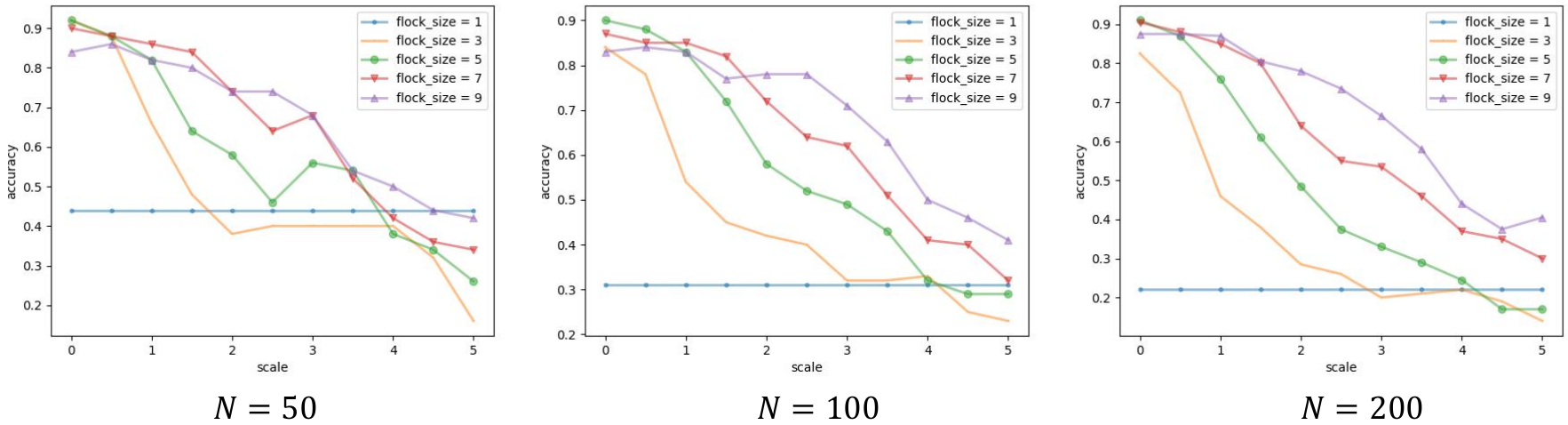}
	\caption{Re-identification accuracy of increasing relative position changes between vehicles (it can be quantified by scale). The curve (flock size$=$1) donates the method based on individual similarity, and curves (flock size$>$1) donate the method based on flock similarity (our method). \textbf{Left:} the number of vehicles is 50. \textbf{Middle:} the number of vehicles is 100.
		\textbf{Right:} the number of vehicles is 200.}\label{fig9}
\end{figure*}

In order to demonstrate the principle of the method based on flock similarity, we visualized the similarity matrix between the two lists, as shown in Fig.~\ref{fig6}. The darker the color, the higher the similarity, and the horizontal and vertical coordinates indicate the position of the individual or flock in two lists. The value on the diagonal represents the similarity between two identical individuals or flocks in the list. When using individual similarity (flock size$=$1), the re-identification accuracy is not high due to the large number of similar individuals in the list. When using flock similarity, the probability of two or more similar flocks appearing in the list is much smaller than the probability of two or more similar individuals appearing in the list, so the application of flock similarity can greatly improve the accuracy of re-identification. The larger the size of the flock, the smaller the probability of two or more similar flocks appearing in the list, and the greater the relative value on the diagonal. However, as the size of the flock grows, the probability of two or more similar individuals appearing within the flock also increases, resulting in a decrease in the accuracy of re-identification. In addition, the increase in flock size also means an increase in the computational complexity of flock similarity. Therefore, the flock size should not be too large.

In order to explore the scope of application of flock similarity, we need to change the relative position of the vehicles as they pass through the two cameras. We only need to adjust the position of the vehicles in List2, i.e., vector $\boldsymbol {y}$.
Let $s_i \sim N(i,\sigma),i=0,1,\ldots,N-1. $
\begin{equation}
	\boldsymbol{s’}=[s_0’,s_1’, \ldots,s_{N-1}’]
\end{equation}
can be obtained by taking a set of samples on $s_i$.
Let $f$ be a reversible mapping satisfying 
\begin{equation}
	s’_{f(0)}<s’_{f(1)}<\ldots<s’_{f(N-1)}
\end{equation}
and $f(i)\in \{0,1,\ldots,N-1\}$, then $y_i=f^{-1}(i)$. We adjust the position of vehicles in List2 according to $y_i$. $y_i=j$ represents that the position in which $x_i$ appears in List2 is $j$.

The larger the scale ($\sigma$) of the normal distribution, the greater the difference in the relative position of the vehicles when they pass through two cameras. Consequently, the scale of the normal distribution can represent the magnitude of the change in the relative position of the vehicles. Therefore, we can adjust the position of vehicles in List2 to varying degrees by using different scales. The results are shown in Fig.~\ref{fig9}.

According to the results, it can be seen that using different scales has no effect on the method based on individual similarity (flock size$=$1). As the scale increases, the relative position between vehicles changes more when passing through two cameras. The effectiveness of the method based on flock similarity gradually decreases, and the rate of decline is related to the size of the flock. Overall, the larger the flock, the slower the decline speed, and the better the robustness to the relative position changes between vehicles when passing through two cameras. This is related to the calculation method of flock similarity.

In fact, we cannot directly obtain the scale of the normal distribution. Therefore, we need another quantifiable metric to measure the change in relative position between vehicles as they pass through two cameras. What we can observe is vector $\boldsymbol {y}$. With $x_i$ as the abscissa and $y_i$ as the ordinate, the relative position change between the vehicles can be visualized.
\begin{figure}[t]
	\centering
	\includegraphics[width=0.35\textwidth]{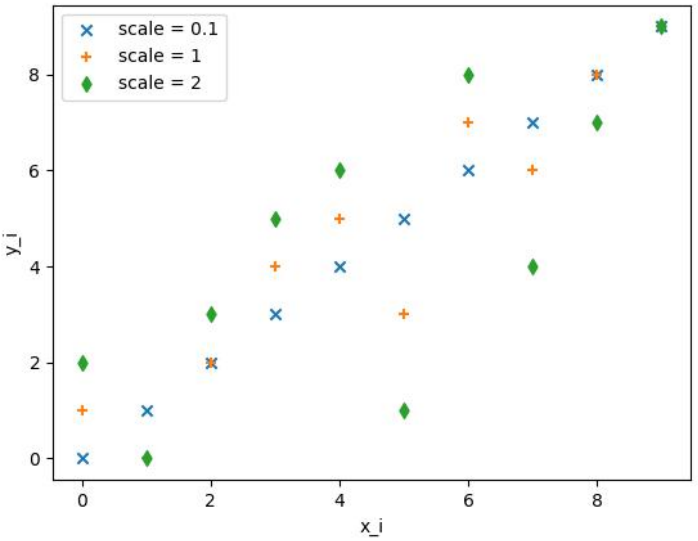}
	\caption{Visualization of the change in relative position between vehicles as they pass through two cameras. $x_i$ represents the order in which $vehicle_i$ appears in Camera1, while $y_i$ represents the order in which $vehicle_i$ appears in Camera2.}\label{fig8}
\end{figure}
As you can see from Fig.~\ref{fig8}, when the scale is small, the scatter distribution is around the line $y=x$, and the larger the scale, the farther the scatter distribution is from the line $y=x$. We can calculate the variance from N scatters $(x_i,y_i)$ to the line $x$=$y$ by the following formula
\begin{equation}
	var=\frac{1}{2n}\sum_{i=1}^{n}(x_i-y_i)^2
\end{equation}
There is a positive correlation between the scale of the normal distribution and the variance from N scatters $(x_i,y_i)$ to the line $y=x$. Fig.~\ref{fig11} shows the corresponding relationship between them.
\begin{figure}[t]
	\centering
	\includegraphics[width=0.35\textwidth]{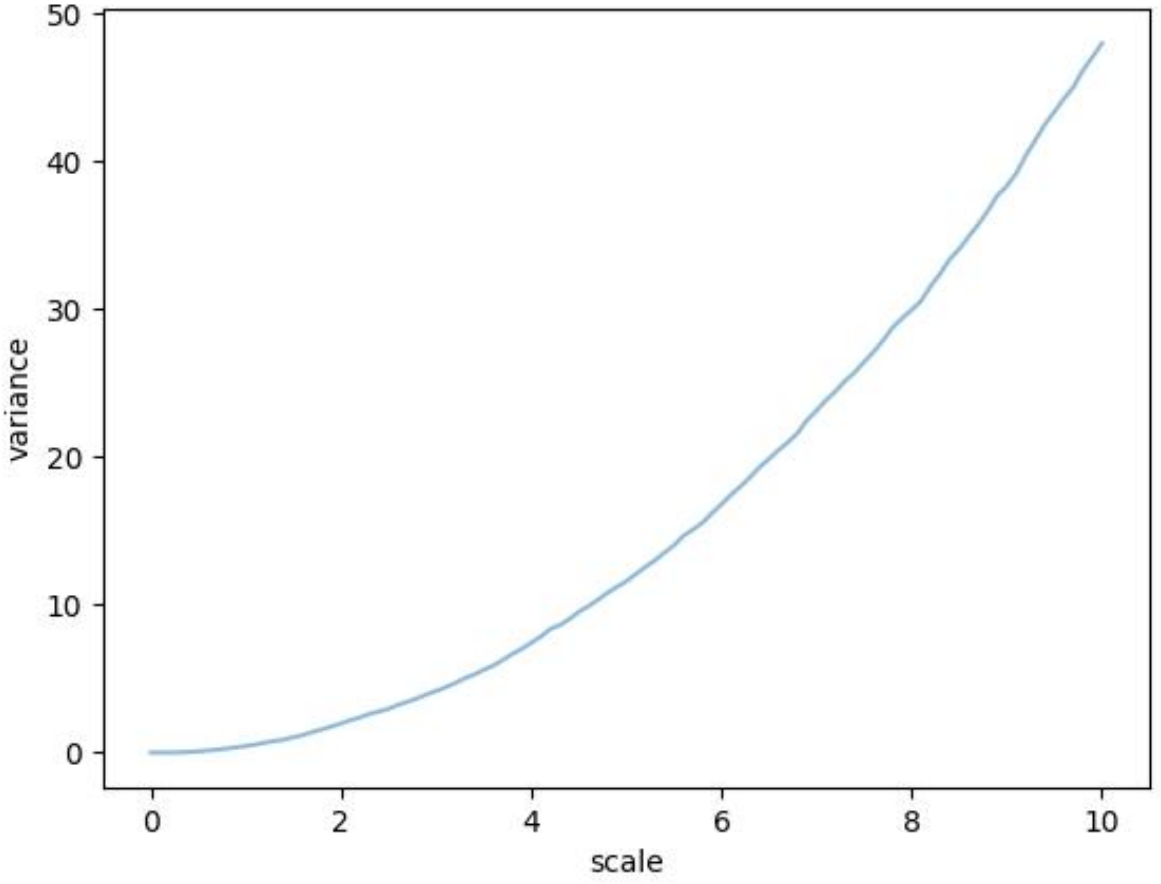}
	\caption{The relationship between two metrics used to measure the change in relative position between vehicles as they pass through two cameras.}\label{fig11}
\end{figure}
In this interval, it can be approximated by the quadratic function $y$=$0.4827x^2+0.01875x-0.0275$. Therefore, the conversion between them can also be approximately calculated using the following formula
\begin{equation}
	scale=1.036\sqrt{ 0.0534+1.93*var}-0.0194
\end{equation}
\section{Conclusion}
In summary, we have introduced a vehicle re-identification method based on flock similarity. When the relative position between vehicles remains unchanged as they pass through two cameras, the accuracy of our method is much higher than the method based on individual similarity (flock size$=$1). In addition to this, our method shows better robustness to the increase in the number of vehicles. The impact of the relative position change between vehicles on the method based on flock similarity is related to the flock size. When the flock size is larger, the robustness of the method is better. However, as the flock size increases, the probability of two or more similar individuals appearing within the flock also increases, leading to a decrease in the accuracy of re-identification. Therefore, the flock size cannot be too large. Although this assumption is based on the bridge scenario, it is often true in other scenarios due to driving safety and camera location.

\section*{Acknowledgment}

This work is supported by the National Natural Science Foundation of China (Grants No. 52372321).

\bibliographystyle{IEEEtran}

\bibliography{your-references}

\begin{thebibliography}{10}
\providecommand{\url}[1]{#1}
\csname url@samestyle\endcsname
\providecommand{\newblock}{\relax}
\providecommand{\bibinfo}[2]{#2}
\providecommand{\BIBentrySTDinterwordspacing}{\spaceskip=0pt\relax}
\providecommand{\BIBentryALTinterwordstretchfactor}{4}
\providecommand{\BIBentryALTinterwordspacing}{\spaceskip=\fontdimen2\font plus
\BIBentryALTinterwordstretchfactor\fontdimen3\font minus
  \fontdimen4\font\relax}
\providecommand{\BIBforeignlanguage}[2]{{%
\expandafter\ifx\csname l@#1\endcsname\relax
\typeout{** WARNING: IEEEtran.bst: No hyphenation pattern has been}%
\typeout{** loaded for the language `#1'. Using the pattern for}%
\typeout{** the default language instead.}%
\else
\language=\csname l@#1\endcsname
\fi
#2}}
\providecommand{\BIBdecl}{\relax}
\BIBdecl

\bibitem{r4}
V.~Eckstein, A.~Schumann, and A.~Specker, ``Large scale vehicle
  re-identification by knowledge transfer from simulated data and temporal
  attention,'' in \emph{2020 IEEE/CVF Conference on Computer Vision and Pattern
  Recognition Workshops (CVPRW)}, 2020, pp. 2626--2631.

\bibitem{r2}
E.~Kamenou, J.~M. del Rincon, P.~Miller, and P.~Devlin-Hill, ``Multi-level deep
  learning vehicle re-identification using ranked-based loss functions,'' in
  \emph{2020 25th International Conference on Pattern Recognition (ICPR)},
  2021, pp. 9099--9106.

\bibitem{r11}
X.~Liu, W.~Liu, T.~Mei, and H.~Ma, ``{PROVID}: Progressive and multimodal
  vehicle reidentification for large-scale urban surveillance,'' \emph{IEEE
  Transactions on Multimedia}, vol.~20, no.~3, pp. 645--658, 2018.

\bibitem{r7}
N.~Peri, P.~Khorramshahi, S.~S. Rambhatla, V.~Shenoy, S.~Rawat, J.-C. Chen, and
  R.~Chellappa, ``Towards real-time systems for vehicle re-identification,
  multi-camera tracking, and anomaly detection,'' in \emph{2020 IEEE/CVF
  Conference on Computer Vision and Pattern Recognition Workshops (CVPRW)},
  2020, pp. 2648--2657.

\bibitem{r9}
Z.~Zheng, T.~Ruan, Y.~Wei, Y.~Yang, and T.~Mei, ``{VehicleNet}: Learning robust
  visual representation for vehicle re-identification,'' \emph{IEEE
  Transactions on Multimedia}, vol.~23, pp. 2683--2693, 2021.

\bibitem{r13}
X.~Ning, K.~Gong, W.~Li, L.~Zhang, X.~Bai, and S.~Tian, ``Feature refinement
  and filter network for person re-identification,'' \emph{IEEE TRANSACTIONS ON
  CIRCUITS AND SYSTEMS FOR VIDEO TECHNOLOGY}, vol.~31, no.~9, pp. 3391--3402,
  SEP. 2021.

\bibitem{r14}
X.~Ning, K.~Gong, W.~Li, and L.~Zhang, ``{JWSAA}: Joint weak saliency and
  attention aware for person re-identification,'' \emph{NEUROCOMPUTING}, vol.
  453, pp. 801--811, SEP. 2021.

\bibitem{r15}
C.~Yan, G.~Pang, X.~Bai, C.~Liu, X.~Ning, L.~Gu, and J.~Zhou, ``Beyond triplet
  loss: Person re-identification with fine-grained difference-aware pairwise
  loss,'' \emph{IEEE TRANSACTIONS ON MULTIMEDIA}, vol.~24, pp. 1665--1677,
  2022.

\bibitem{r16}
T.~Si, F.~He, Z.~Zhang, and Y.~Duan, ``Hybrid contrastive learning for
  unsupervised person re-identification,'' \emph{IEEE TRANSACTIONS ON
  MULTIMEDIA}, vol.~25, pp. 4323--4334, 2023.

\bibitem{r1}
N.~Jiang, Y.~Xu, Z.~Zhou, and W.~Wu, ``Multi-attribute driven vehicle
  re-identification with spatial-temporal re-ranking,'' in \emph{2018 25th IEEE
  International Conference on Image Processing (ICIP)}, 2018, pp. 858--862.

\bibitem{r3}
X.~Chen, H.~Sui, J.~Fang, W.~Feng, and M.~Zhou, ``Vehicle re-identification
  using distance-based global and partial multi-regional feature learning,''
  \emph{IEEE Transactions on Intelligent Transportation Systems}, vol.~22,
  no.~2, pp. 1276--1286, 2021.

\bibitem{r8}
S.~Jaiswal, P.~Chakraborty, T.~Huang, and A.~Sharma, ``Traffic intersection
  vehicle movement counts with temporal and visual similarity based
  re-identification,'' in \emph{2023 8th International Conference on Models and
  Technologies for Intelligent Transportation Systems (MT-ITS)}, 2023, pp.
  1--6.

\bibitem{r5}
Y.~Chen, S.~Zhang, F.~Liu, C.~Wu, K.~Guo, and Z.~Qi, ``{DVHN}: A deep hashing
  framework for large-scale vehicle re-identification,'' \emph{IEEE
  Transactions on Intelligent Transportation Systems}, vol.~24, no.~9, pp.
  9268--9280, 2023.

\bibitem{r6}
P.~Khorramshahi, V.~Shenoy, and R.~Chellappa, ``Robust and scalable vehicle
  re-identification via self-supervision,'' in \emph{2023 IEEE/CVF Conference
  on Computer Vision and Pattern Recognition Workshops (CVPRW)}, 2023, pp.
  5295--5304.

\bibitem{r18}
G.~Yang, P.~Wang, W.~Han, S.~Chen, S.~Zhang, and Y.~Yuan, ``Automatic
  generation of fine-grained traffic load spectrum via fusion of
  weigh-in-motion and vehicle spatial-temporal information,''
  \emph{COMPUTER-AIDED CIVIL AND INFRASTRUCTURE ENGINEERING}, vol.~37, no.~4,
  pp. 485--499, MAR. 2022.

\bibitem{r12}
K.~Simonyan and A.~Zisserman, ``Very deep convolutional networks for
  large-scale image recognition,'' \emph{arXiv 1409.1556}, SEP. 2014.

\bibitem{r10}
X.~Liu, W.~Liu, H.~Ma, and H.~Fu, ``Large-scale vehicle re-identification in
  urban surveillance videos,'' in \emph{2016 IEEE International Conference on
  Multimedia and Expo (ICME)}, 2016, pp. 1--6.

\end{thebibliography}
\end{document}